\documentclass[runningheads]{llncs}
\usepackage{graphicx}

\usepackage{tikz}
\usepackage{comment}
\usepackage{amsmath,amssymb} 
\usepackage{color}

\usepackage[activate={true,nocompatibility},final,tracking=true,kerning=true,spacing=true,factor=1100,stretch=10,shrink=9]{microtype}
\usepackage{times}
\usepackage{epsfig}
\usepackage{graphicx}
\usepackage{amsmath}
\usepackage{amssymb}
\usepackage{adjustbox}
\usepackage{float}
\usepackage{pifont}
\newcommand{\cmark}{\ding{51}}%
\newcommand{\xmark}{\ding{55}}%

\DeclareMathOperator{\E}{\mathbb{E}}

\usepackage[pagebackref=true,breaklinks=true,colorlinks,bookmarks=false]{hyperref}
\renewcommand*\backref[1]{\ifx#1\relax \else (Cited on page #1) \fi}

\usepackage{xcolor}
\definecolor{RED}{rgb}{1, 0, 0}
\definecolor{fig_red}{rgb}{1, 0, 0}
\definecolor{fig_green}{rgb}{0, 0.6, 0}
\newcommand{\squeezeup}{\vspace{-2mm}}

\begin{document}
\pagestyle{headings}
\mainmatter
\def\ECCVSubNumber{9}  

\title{HyperNST: Hyper-Networks for Neural Style Transfer} 

\titlerunning{HyperNST: Hyper-Networks for Neural Style Transfer}
%
\author{Dan Ruta\inst{1} \and
Andrew Gilbert\inst{1} \and
Saeid Motiian\inst{2} \and
Baldo Faieta\inst{2} \and
Zhe Lin\inst{2} \and
John Collomosse\inst{1,2}
}%
\authorrunning{D. Ruta et al.}
%
\institute{University of Surrey \and
Adobe Research 
}
\maketitle

\begin{figure}[h]
    \centering
    \includegraphics[width=0.9\linewidth]{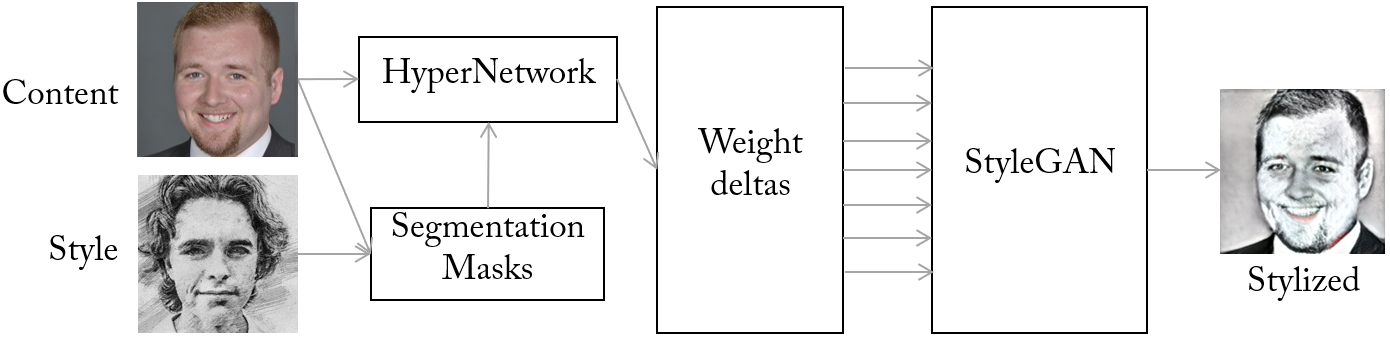} 
    \label{fig:teaser}
    \squeezeup
    \squeezeup
    \squeezeup
\end{figure}

\begin{abstract}

We present HyperNST; a  neural style transfer (NST) technique for the artistic stylization of images, based on Hyper-networks and the StyleGAN2 architecture.   Our contribution is a novel method for inducing style transfer parameterized by a metric space, pre-trained for style-based visual search (SBVS).   We show for the first time that such space may be used to drive NST, enabling the application and interpolation of styles from an SBVS system. The technical contribution is a hyper-network that predicts weight updates to a StyleGAN2 pre-trained over a diverse gamut of artistic content (portraits), tailoring the style parameterization on a per-region basis using a semantic map of the facial regions.  We show HyperNST to exceed state of the art in content preservation for our stylized content while retaining good style transfer performance.


\end{abstract}

\section{Introduction}

Neural style transfer (NST) methods seek to transform an image to emulate a given appearance or `style' while holding the content or structure unchanged.  Efficient, stylistically diverse NST remains an open challenge. Feed-forward NST methods are fast yet typically fail to represent a rich gamut of styles. At the same time, optimization based approaches can take several seconds or minutes at inference time, lacking the speed for practical, creative use.  Moreover, NST algorithms are often driven by one or more exemplar `style' images, rather than an intuitive parameter space,  impacting their controllability as a creative tool.

In this work, we propose a fast feed-forward method for driving neural stylization (NST) parameterized by a metric embedding for style representation.  Our approach is based upon a hyper-network trained to emit weight updates to a StyleGAN2\cite{stylegan2} model, trained on a large dataset of artistic portraits (e.g. AAHQ \cite{aahq}) in order to specialize it to the depiction of the given target style.  Our work is inspired by the recent StyleGAN-NADA \cite{stylegan_nada}, in which a CLIP \cite{clip} objective is optimized by fine-tuning a pre-trained StyleGAN2 model to induce NST.  We extend this concept to a feed-forward framework using a hyper-network to generate the weight updates.   Furthermore, we introduce the use of a metric parameter space   (ALADIN \cite{aladin}) originally proposed for style driven visual search to condition the hyper-network prediction (vs. CLIP in \cite{stylegan_nada}) and adaptively drive this parameterization using a semantic map derived from the source and target image.  Without loss of generality, our experiments focus on the challenging domain of facial portraits driven using a semantic segmentation algorithm for this content class.  We show our method improves target image content retention versus the state of the art while retaining comparable accuracy for diverse style transfer from a single model -- and despite using a hyper-network exhibiting comparable inference speed to leading feed-forward NST  \cite{swapping_ae}.   Moreover, our method exhibits good controllability; using a metric space for our style code enables intuitive interpolation between diverse styles and region-level controllability of those parameters.  We adopt the recent ALADIN style code for this purpose, raising the novel direction of unifying style based visual search and stylization from a single representation.

\section{Related Work}

\noindent{\bf Neural Style Transfer (NST)} The seminal work of Gatys et al. \cite{gatys} enabled artistic style transfer through neural models. This work demonstrates the correlation between artistic style and features extracted from specific layers in a pre-trained vision model.
The AdaIN work \cite{HuangAdaIn2017} introduced parameterized style transfer through first and second moment matching, via mean and standard deviation values extracted from random target style images.
MUNIT \cite{munit} explores domain translation in images through the de-construction of images into semantic content maps and global appearance codes.
ALADIN \cite{aladin} explored the creation of a metric space modeling artistic style, across numerous areas of style. The embedding space was trained in conjunction with AdaIN and a modified version of the MUNIT model and weakly supervised contrastive learning over the BAM-FG dataset.
A follow-up work \cite{stylebabel} studying multi-modal artistic style representation expanded upon ALADIN, pushing the representation quality further through a vision transformer model.
CycleGAN also explored domain transfer in images, but through learned model weight spaces, encoding pairwise image translation functions into separate generators for each image domain. Enforced by cyclic consistency, the translation quality between a pair of image domains was high, at the cost of requiring bespoke models for each domain translation to be trained.
Using StyleGAN as a generator model, Swapping Autoencoders \cite{swapping_ae} directly learn the embedding of images into a StyleGAN generator's weight space while simultaneously encoding a vector representation of the visual appearance of an image externally. These models separately focus on landscapes, buildings, bedrooms, or faces.\\

\noindent{\bf StyleGAN Inversion}  The evolution of the StyleGAN models \cite{styleGAN,stylegan2,stylegan3} explore generation of extremely realistic portrait images. They use weight modulation based editing of visual appearance in the generated images. They also undertake preliminary investigations into the inversion of existing images into the GANs' weight spaces for reconstruction. The work in e4e \cite{e4e} includes an undertaking of a deeper analysis of real image embedding into the StyleGAN weight space, including the quality/editability tradeoff this imposes. Their work enables multiple vectors of editability for images generated through StyleGAN across several domains.  Restyle \cite{restyle} improves the reconstruction quality of images embedded into the weight space by executing three fine-tuning optimization steps at run-time. Also, using StyleGAN as target generators, HyperStyle \cite{hyperstyle} embeds real portrait images into the weight space with high fidelity. Similar to our approach, they use a hyper-network to generate weight updates for StyleGAN, trained to infer weights to bridge the gap between quick rough inversions and the fully detailed reference portrait images. Strengths of this approach are the high reconstruction fidelity and strong photorealistic editing control for portrait photos. They further undertake some early explorations at domain adaptation for images by changing StyleGAN checkpoints. However, the HyperStyle work focuses on photorealism and does not enable region-based control or style space parameterization as we propose in HyperNST.

Our approach is inspired by the StyleGAN-NADA \cite{stylegan_nada} work, which explores style transfer in the StyleGAN weight space through CLIP-based optimization. Though effective, this incurs long-running optimization passes, which are impractical for wide use. Moreover, this method has no built-in methods to effectively embed real portrait images into the weight space for reconstruction and editing.
Recently, FaRL \cite{farl_pyfacer} undertakes representation learning of facial images, through multi-modal vision-text data, with face segmentation models that cope well with various visual appearances for portraits.



\section{Methodology}

Domain adaptation models like CycleGAN perform well at image translation, where style features from a style image are correctly mapped to the matching semantic features in a content image. A limitation of CycleGAN is that transfer only between a single pair of image domains (styles) is possible.  Hypernetwork \cite{hypernetwork} models are used to predict the weights of a target model. Such a hypernetwork can be conditioned on some input before inferring the target model weights.  Early experiments with a CycleGAN backbone yielded promising results, especially when we re-framed the training process into a more manageable task by learning weight offsets to a pre-trained checkpoint rather than a from-scratch model. Motivated by these findings, we use a more extensive, modern generator such as StyleGAN2 with weight updates predicted using a hyper-network.  This echoes the recent optimization based approach StyleGAN-NADA which updates weights via fine-tuning for stylization rather than hyper-network prediction. HyperStyle \cite{hyperstyle} uses a channel-wise mean shift in the target generator, which significantly reduces the target number of trainable weights into a practical range.

\subsection{HyperNST Architecture}
\begin{figure}[t!]
    \centering
    \includegraphics[width=\linewidth]{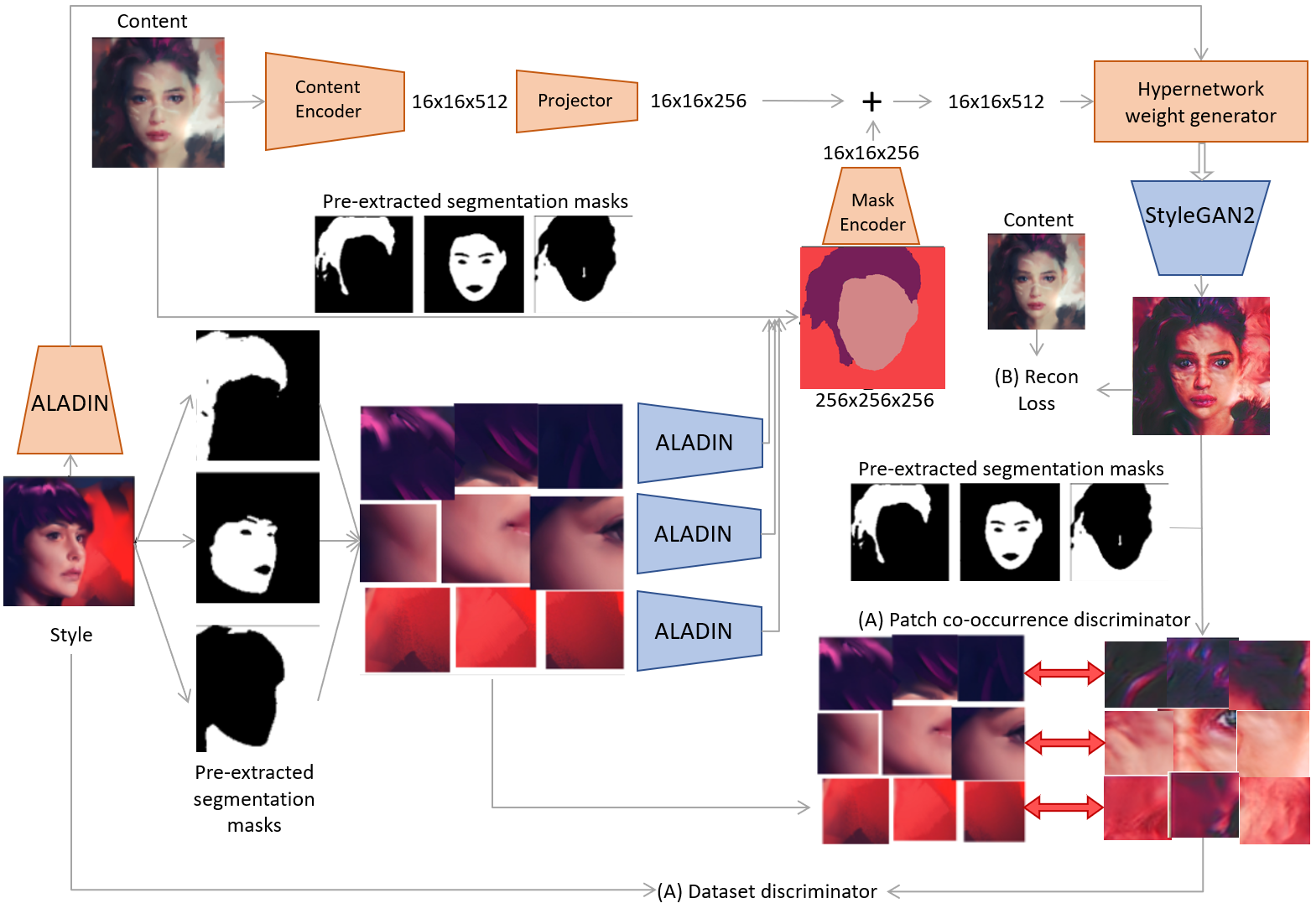} 
    \caption{Architecture diagram of our approach. Facial semantic segmentation regions used in conditioning via ALADIN and guiding via patch co-occurence discriminator a HyperStyle model into embedding a content portrait image into an AAHQ+FFHQ trained StyleGAN2 model, and using ALADIN style codes to stylize it towards the style of a style image.   
    Blue modules represent frozen modules, and orange modules represent modules included in the training. + represents concatenation in the channels dimension.
    (A) represents the stylization pass losses, and (B) represents the reconstruction pass losses. Not pictured for clarity: the reconstruction pass uses the same semantic regions as the content image for the 256x256x256 semantically arranged ALADIN conditioning. 
    }
    \label{fig:arch_diagram}
    \squeezeup
    \squeezeup
    
\end{figure}

We compose the stylization pipeline of our model with a hyper-network set up to generate weight deltas for a target frozen, pre-trained StyleGAN2 model. Fig \ref{fig:arch_diagram} shows an architecture diagram of HyperNST, showing the losses and the conditioning process of the hyper-network upon the content image and semantically arranged ALADIN style codes of the style image.

We direct the training to find a set of weight deltas for StyleGAN2, which can generate the same image as the content image for reconstruction. We begin with the GAN inversion, using e4e to reconstruct a given content portrait image roughly. An encoder encodes this image into a tensor used by the hyper-network weight generator modules along with style information to predict weight deltas to apply to the frozen, pre-trained StyleGAN2. 
Stylization then occurs by changing the style embedding in the hypernetwork conditioning.

\subsection{Conditioning on style, and facial semantic regions}

We further introduce conditioning on a target style image by injecting a style representation embedding. We project the initial 16x16x512 encoding into 16x16x256, using ALADIN \cite{aladin} codes to compose the other half of the 16x16x512 tensor. Each ALADIN code is a 1x256 vector representing the artistic style of an image. We project semantically arranged ALADIN embeddings into a second 16x16x256, which we then concatenate together with the first to form the 16x16x512 tensor upon which the hypernetwork is conditioned.

Facial portraits are very heavily grounded in their semantic structure. We therefore condition the stylization process on these semantic regions to ensure that the content is maintained. We execute this using segmentation masks extracted via FaRL \cite{farl_pyfacer}. 

To condition on semantic regions, we aim to use the ALADIN style code representing only the style contained within a given semantic region. Given that ALADIN codes can only be extracted from square or rectangular inputs, we use the average ALADIN code extracted across several patches per semantic region.

We use FaRL to compute the semantic regions, extract patches from the image randomly, and use mean intersection over union (IoU) to ensure that patches mainly cover pixels attributed to these respective semantic classes for each region. The ALADIN codes from these patches are averaged to form the average ALADIN code for a given semantic class. These average style codes are tiled and re-arranged on a per-pixel basis to match the semantic segmentation maps of the \textit{content} image before they are projected by an encoder into the second 16x16x256 tensor, as above.

This projected 16x16x256 semantically arranged ALADIN tensor and the original 16x16x256 tensor encoded from the content image, are concatenated in the channel dimension to 16x16x512, making up the final tensor which the hypernetwork weight delta generating modules operate over.

\subsection{HyperNST training process}

We use three iterations per batch refining methodology during training. We include a patch co-occurrence discriminator $D_{patch}$ to introduce a style loss and an image-level discriminator $D$. The hypernetwork weight deltas generator $H$ is conditioned on the encoded content images $E_c(c)$, and ALADIN style codes arranged by the semantic segmentation mask of the content image $M(x, E_{mask})$, seen in Eq \ref{eq:mask}.

\begin{equation}
    \label{eq:mask}
    M(x, E_{mask}) = E_{mask}\left( rearrange \left( \frac{\sum_{p=1}^{P} A(crop(x, SM_x)_p)}{P}, SM_c  \right) \right)
\end{equation}

\begin{equation}
    \label{eq:stylization}
    y(c,s) = G(\theta + H(M(s) ^\frown E_{c}(c), A(s))))
\end{equation}

During training, two forward passes are executed, one for stylization (A) (Eq \ref{eq:stylization}), and one for reconstruction (B). 
The reconstruction pass (B) loads content images and their respective ALADIN style code arranged by the content image semantic map and performs the original reconstruction code (Eq. \ref{eq:recon}). 
The stylization pass (A) performs style transfer by loading the content images $c$, and ALADIN style codes of other style images $s$. During stylization, the generation is performed with the mixed features combined to create stylized images, with the patch co-occurrence discriminator providing a learning signal to train the stylization (Eq. \ref{eq:pcd}). In this pass, the discriminator is also trained, with the stylized images as \textit{fake}, and the target style images as \textit{real} (Eq \ref{eq:disc}).

\begin{equation}
\label{eq:recon}
\mathcal{L}_{rec}(c) =  \left( c, y(c,c)) \right) 
\end{equation}

\begin{multline}
    \label{eq:pcd}
    \mathcal{L}_{pcd}(D_{patch}) =  
    \E_{s \sim S, c \sim C}  \left[ -log(D_{patch}(crops(s), crops(y(c,s)))) \right]
\end{multline}

\begin{equation}
    \label{eq:disc}
    \mathcal{L}_{disc}(c, s, D) = \E_{s \sim S, c \sim C}  \left[ -log(D(s, y(c,s)) \right]
\end{equation}

where: $c$ is content image, $s$ is style image, $C$ are all content images, $S$ are all style images, $\theta$ are the original StyleGAN2 weights, $H$ is the hypernetwork weight delta generator, $G$ is the StyleGAN2 generator, $A$ is ALADIN, $P$ is number of patches per semantic region, $SM$ are pre-extracted semantic segmentation masks, and $^\frown$ represents concatenation. 



\begin{equation}
    \label{eq:final}
    \mathcal{L} = \lambda_{1} \mathcal{L}_{rec} + \lambda_{2} ( \mathcal{L}_{disc} +  \mathcal{L}_{pcd} )
\end{equation}

The final loss is shown in Eq. \ref{eq:final}, with Sec.\ref{sec:ablations} describing ablations for the $\lambda$ values.


\subsection{Stylized target generator}

The hypernetwork model is trained to infer weight delta values for weights in a StyleGAN2 model, which acts as the generator. Given that we work in the artistic style domain, we thus need the target generator to be able to model the weight space of a generator already able to produce high quality highly stylized images. We thus first fine-tune an FFHQ \cite{ffhq} StyleGAN2 model on the Artstation Artistic Face (AAHQ) dataset \cite{aahq}. Largely popular in research centered around the facial images domain, the FFHQ dataset has been used by the majority of reference papers targeting this domain. Given our exploration into artistically manipulating the visual features of portrait images, we further use this AAHQ dataset, which encompasses portraits from across a large and varied corpus of artistic media. We continue to include FFHQ images, to ensure we keep high quality modeling capabilities for features more often found in photographic images than artistic renderings (like glasses and beards) to ensure real world images can still be encoded well. 

We first train our model simply for ALADIN conditioned reconstruction of the AAHQ and FFHQ datasets for all layers, including the toRGB layers, which have been shown to target textures and colors \cite{hyperstyle}. We then fine-tune this pre-trained checkpoint with the goal of stylization, where we include the patch co-occurrence training for guidance and ALADIN code swapping in conditioning for style transfer.

During the stylization fine-tuning step, we freeze the content encoder and train only the hypernetwork weight delta prediction modules. This is to prevent the stylization signal from negatively affecting the model's embedding abilities to reconstruct real images with high accuracy.

We run the stylization fine-tuning only on the part of the target StyleGAN layers. We find that the further into the StyleGAN2 model we apply stylization fine-tuning, the more the stylization affects colorization rather than textures and adjustments to the structure. We include the toRGB layers originally omitted in HyperStyle for their texture adjustments, and we train the weight deltas generation modules for layers 13 onward (out of 25). The weight delta generation for layers before this are frozen after their initial training, therefore still allowing reconstruction, but no more extended training during the stylization stage. This ensures that the overall facial structure of the images is not too greatly affected during stylization training. Layer 13 is a \textit{sweet middle spot} with the best balance between stylization and retaining good face structure reconstruction. A visual example of this phenomenon can be seen in Fig.\ref{fig:layers_viz}.

We further make changes to the patch co-occurrence discriminator. The Swapping Autoencoders model generates images in 1024x1024 resolution, with a discriminator operating over 8 patches of 128x128 dimensions. Our hypernetwork model generates images with a resolution of 256x256 pixels, therefore, we adjust our discriminator's patch sizes to a lower size of 32x32.

\subsection{Region mask driven patch discriminator}



We also use facial semantic segmentation masks in the patch co-occurrence discriminator, to provide a style signal separated by semantic region.

In the original discriminator, patches are extracted from across the entire image at random. Instead, our process is repeated for each semantic class' map. Patches are extracted at random, ensuring that the mean intersection over union (IoU) mostly covers pixels attributed to the respective semantic class.  As in the original case, the losses from the patches are averaged to form the style learning signal. However, instead of comparing the stylized patches with patches randomly selected from the reference style image, we select patches bound by the respective semantic region - e.g. style hair patches are matched with stylized hair patches, background with background, face with face, etc. 


\subsubsection{Implementation}

The semantic regions predicted by FaRL contain several regions which individually cover only small areas in the pixel space. Given our use of the regions as patches we extract for ALADIN and the patch co-occurrence discriminator, we could not accurately use these small regions without the patches mainly containing pixel data from the surrounding regions. Instead, we group up the regions into 3 larger classes which typically cover larger areas of a portrait: (1) Hair, (2) Face, and (3) Background. Due to overlap that cannot be avoided, the \textit{Face} region also contains eyes, noses, and lips. Furthermore, some images do not contain hair, for which this semantic class is not used. In the semantically driven patch co-occurrence loss, we use 3 patches for each region, totaling 9 patches.


\section{Evaluation}

Experiments generally required around 24 hours to converge. We ran our experiments on a single NVIDIA RTX 3090 with 24GB of VRAM. The HyperNST experiments were executed with a batch size of 1, which required around 22GB of VRAM, due to the high number of weights needed in the hypernetwork configuration. 

\subsection{Datasets}

We create a test dataset for use in evaluation, using images from FFHQ \cite{ffhq}, and AAHQ \cite{aahq}. We extract 100 random content images from FFHQ and 100 random style images from AAHQ. Together, these result in 10'000 evaluation images, when models are used to stylize all combinations.

We measure the content similarity using the two LPIPS variants (Alexnet \cite{alexnet}, and VGG \cite{vgg}), and we measure the style similarity using SIFID.


\textbf{LPIPS} \cite{lpips} evaluates the variance between two images based on perceptual information better aligned to human perception compared to more traditional statistics based methods. We use this LPIPS variation to compute the average variation between each stylized image and its original content image. A lower average variation value would indicate a more similar semantic structure, therefore, a lower value is better.


Like Swapping Autoencoders, we employ the Single Image Fréchet Inception Distance (\textbf{SIFID}) introduced in SinGAN \cite{singan}. This metric evaluates FID, but uses only a single image at a time, which is most appropriate when evaluating style transfer computed using a single source sample. A lower value indicates better style transfer here. 

Finally, we measure the time it takes to synthesize the stylized image in seconds per image. A low inference time is essential for the practicality of a model in real use cases. All timings were computed on an RTX 3090 GPU.


\subsection{Baselines}


To test the performance of our approach, we train 7 other methods on the FFHQ and AAHQ datasets: Swapping Autoencoders \cite{swapping_ae}, Gatys \cite{gatys}, ArtFlow \cite{artflow}, PAMA \cite{pama}, SANet \cite{sanet}, NNST \cite{nnst}, and ContraAST \cite{contraAST}. From these, Swapping Autoencoders is the most closely related, as it also operates over a StyleGAN2 space, thus, we separate it from the other more traditional methods which do not.

Table \ref{tab:ablations_table} shows the LPIPS, SIFID, and timing results computed for our own HyperNST method, 
and the other methods we evaluate against.
The results show the superiority of HyperNST in retaining the most semantic structure in the facial reconstructions, indicated by the lowest LPIPS value. Meanwhile, we achieve comparable SIFID values (2.279) to Swapping AutoEncoders (1.948), indicating a small gap in stylization quality. Only our HyperNST model can embed a portrait into a model with semantic editing capabilities. 
Finally, the inference time of our model is similar to other methods and low enough for practical use of the model in real applications.

One disadvantage of the hypernetwork based approach is the current limitation with how strongly the StyleGAN2 model can be fine-tuned. As per our findings and the HyperStyle authors', hyper-learning weight deltas for every individual weight would be infeasible on current hardware (requiring over 3 billion parameters for a StyleGAN2 model) - we are limited by the technology of our time. So at best, we can aim to at most match a method where the model is fully trained, the closest comparison being Swapping AutoEncoders. The Swapping AutoEncoders model also fine-tunes a StyleGAN2 model. However, the more straightforward training approach affords the luxury to fine-tune individual weights in the generator, compared to the channel-wise mean shifts, currently possible with a hyper-network on today's hardware.

\begin{table}[t!]
  \centering
  \small
      \centering
      \begin{tabular}{l|c|c|c|c|c|c}
        \hline
        Model & LPIPS (Alexnet) & LPIPS (VGG) & SIFID & Time (s/img) & Interpolation & Editing \\
        \hline

        SAE \cite{swapping_ae} & 0.334500 & 0.4857 & 1.948 & 0.10 & \cmark & \xmark \\ 

        HyperNST (ours) & 0.000042 & 0.0017 & 2.279 & 0.35 & \cmark & \cmark \\ 
        \hline
        
        Gatys \cite{gatys} & 0.000164 & 0.0030 & 1.369 & 14.43 & \xmark & \xmark \\ 
        
        ArtFlow \cite{artflow} & 0.000080 & 0.0022 & 1.347 & 0.32 & \xmark & \xmark \\ 

        PAMA \cite{pama} & 0.000109 & 0.0029 & 0.522 & 0.14  & \cmark  & \xmark \\ 

        SANet \cite{sanet} & 0.000180 & 0.0068 & 0.486 & 0.11  & \cmark  & \xmark  \\ 
        NNST \cite{nnst} & 0.000149 & 0.0030  & 0.871 & 55.40  & \xmark  & \xmark  \\ 

        ContraAST \cite{contraAST} & 0.000133 & 0.0035 & 0.666 & 0.10  & \xmark  & \xmark  \\ 

        \hline
      \end{tabular}

    \caption{
    Overall results,
    comparing metrics for our model's content, style, and timing and the closest most similar model, Swapping AutoEncoders (top), and some further, more traditional methods (bottom). All methods were trained using AAHQ as the style dataset and FFHQ as the content dataset.
    }
    \label{tab:ablations_table}
  \end{table}
  
Nevertheless, hyper-network based NST is interesting to study, as such limitations are temporary. Both model and hardware improvements with subsequent works will reduce such limitations in time. We can compare hyper-network approaches directly with fully-trained counterparts for the current state of the art. 

\begin{figure}[t!]
    \centering
    \includegraphics[width=0.9\linewidth]{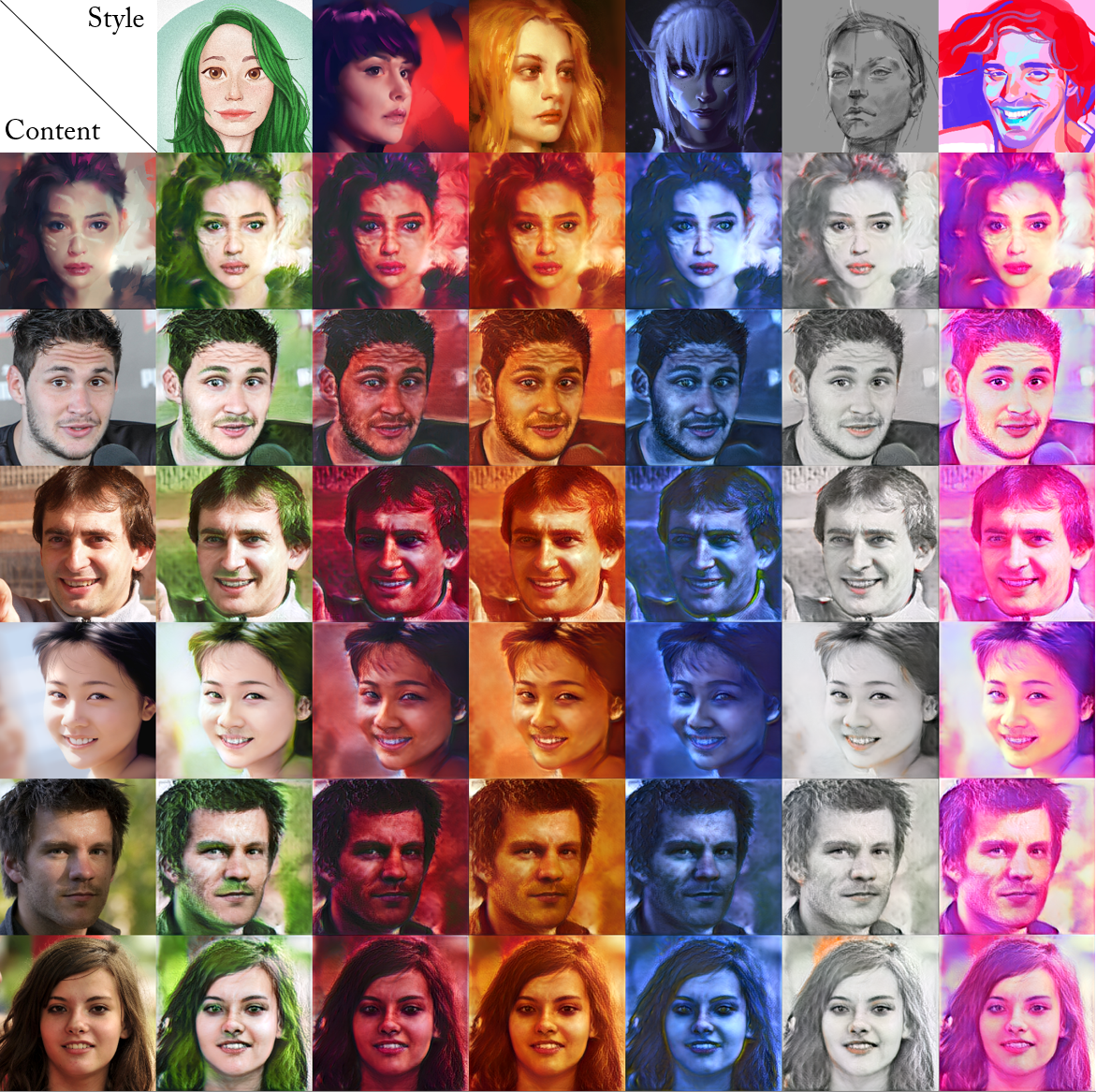} 
    \caption{Qualitative HyperNST visualization of portrait style transfer.}
    \label{fig:hypernst_viz}
    \squeezeup
\end{figure}

\begin{figure}[t!]
    \centering
    \includegraphics[width=0.9\linewidth]{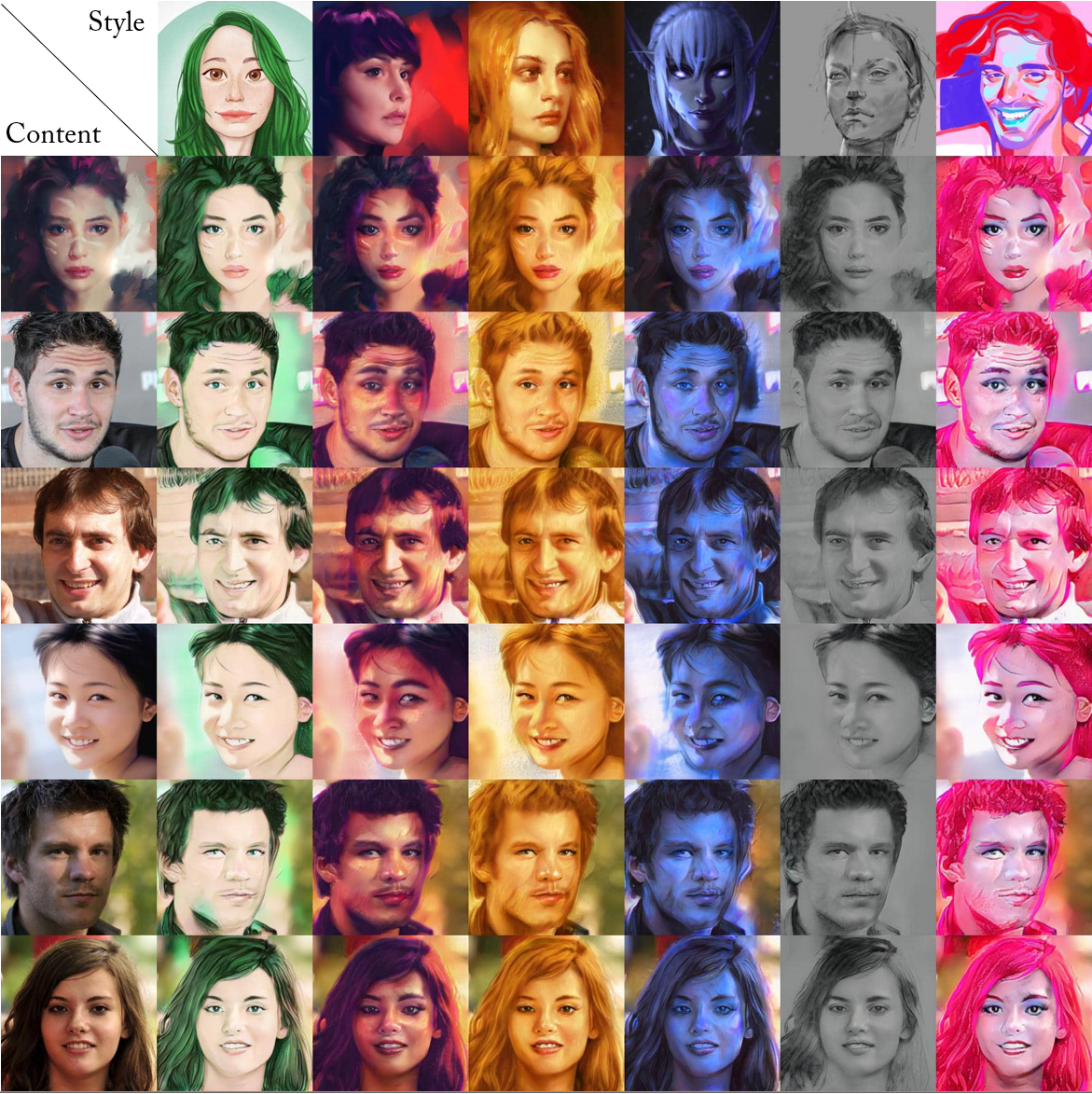} 
    \caption{Baseline comparison to Swapping Autoencoders (re-trained with AAHQ), the closest work - a fully per-weight-trained model.}
    \label{fig:sae_baseline}
    \squeezeup
\end{figure}

Figure \ref{fig:sae_baseline} visualizes some style transfer results obtained with a Swapping Autoencoders (SAE) model, re-trained with the AAHQ dataset. Figure \ref{fig:hypernst_viz} shows the same visualization, obtained with HyperStyle. It is worth noting that the SAE model generates 1024x1024 images, whereas the target StyleGAN2 model in the hypernetwork setting uses a smaller, 256x256 version. Quality of stylization is somewhat equivocal for these approaches; in most cases, the SAE results are slightly more faithful in their style transfer accuracy but are less faithful at retaining the semantic structure of the portraits vs. HyperNST. The SAE model also  has difficulty stylizing backgrounds.


\begin{table}[t!]
  \centering
  \small
      \centering
      \begin{tabular}{l|c|c|c}
        \hline
        Style loss strength & LPIPS (Alexnet) & LPIPS (VGG) & SIFID \\
        \hline
        
        0.5 & 0.001480 & 0.002657 & 3.312  \\ 
        1.0 & 0.001310 & 0.002626 & 2.590  \\ 
        2.0 & 0.000042  & 0.001688 & 2.279 \\ 
        5.0 & 0.000097 & 0.000191 & 2.463  \\ 
        10.0 &0.000091 & 0.000222 & 2.571  \\ 

        \hline
      \end{tabular}

    \caption{Ablation results for varying the style loss strength}
    \label{tab:ablation_style_strength}
  \squeezeup
  \squeezeup
  \squeezeup
\end{table}

\begin{figure}
    \centering
    \includegraphics[width=0.9\linewidth]{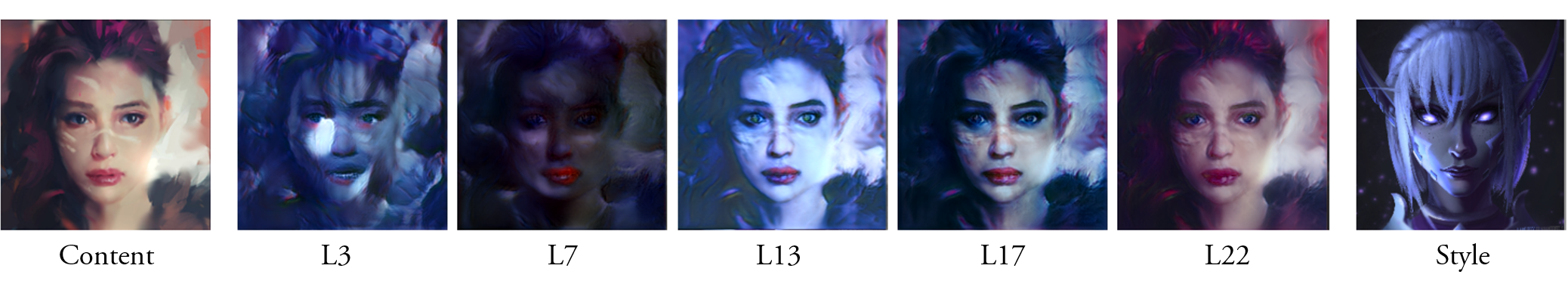} 
    \caption{Representative visualization of the effect of choosing where in the target StyleGAN2 layers to start fine-tuning hyper-weight generators for. The left-most image is the reference content image, and the right-most is the reference style image. In the middle, from left to right, images represent visualizations for layers 3, 7, 13, 17, and 22.}
    \label{fig:layers_viz}
    \squeezeup
\end{figure}


\begin{table}[htbp]
  \centering
  \small
      \centering
      \begin{tabular}{l|c|c|c}
        \hline
        Layer & LPIPS (Alexnet) & LPIPS (VGG) & SIFID \\
        \hline
        
        3 & 0.000140 & 0.00399 & 3.123  \\ 
        7 & 0.000190 & 0.00302 & 2.792 \\ 
        13 & 0.000042  & 0.00169 & 2.279  \\ 
        17 & 0.000153 & 0.00268 & 3.337  \\ 
        22 & 0.000042 & 0.00104 & 3.506  \\ 

        \hline
      \end{tabular}

    \caption{Ablation results for varying the starting StyleGAN2 layer past which layers are included in the stylization fine-tuning.}
    \label{tab:ablation_layers}
  \end{table}
  
\subsection{Ablations}
\label{sec:ablations}

\begin{figure}[]
    \centering
    \includegraphics[width=0.75\linewidth,height=18cm]{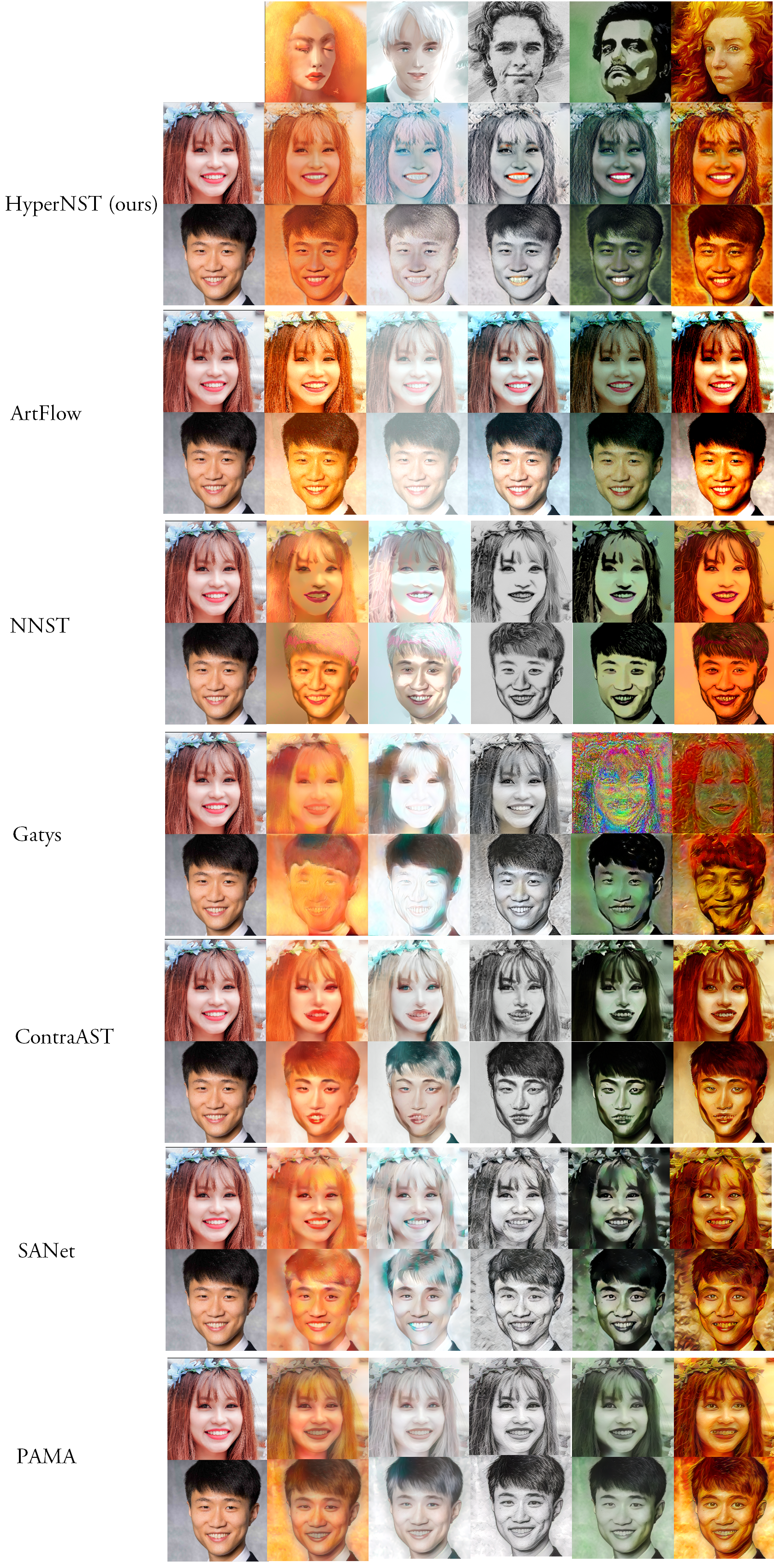} 
    \caption{Stylization comparison of HyperNST (1st pair), against Artflow (2nd), NNST (3rd), Gatys et.al (4th), ContraAST (4th), SANet (5th), and PAMA (6th)}
    \label{fig:nst_comp}
    \squeezeup
\end{figure}





We perform the stylization fine-tuning on only part of the StyleGAN2 layers' hyper-weight generators. Out of 25 layers, we find layer 13 to be a good balance between inducing style transfer and retaining semantic structure. The earlier the target StyleGAN2 layers that hyper-weight generators get fine-tuned for stylization, the stronger the emphasis on style is, therefore losing structure reconstruction quality. Conversely, the later the starting layer is, the less pronounced the stylization is, thereby focusing the training more on the semantic reconstruction quality and instead just performing a color transfer. Figure \ref{fig:layers_viz} shows a representative visualization of this effect, and Tbl. \ref{tab:ablation_layers} shows quantitative metrics. The images further to the left represent a deeper fine-tuning of layers of style and losing structure. The images to the right show a more shallow fine-tuning for style, retaining more structure from the original content image. We targeted these specific layers, as these are toRGB layers in the target StyleGAN2 layer, which are known to more significantly affect color and texture.


Finally, we ablate the use of facial semantic region information in the model and training pipeline. We explore the effect of including the semantic regions as a conditioning signal in the pipeline (cond) and as a way to match region types in the patch co-occurrence discriminator (loss). We populate the results in Tbl\ref{tab:ablation_sem}, showing the usefulness of these components.

\begin{table}[htbp]
  \centering
  \small
      \centering
      \begin{tabular}{l|c|c|c}
        \hline
        Experiment & LPIPS (Alexnet) & LPIPS (VGG) & SIFID \\
        \hline
        Baseline (with both cond and loss) & 0.00004 & 0.0017 & 2.279  \\ 
        \hline
        No mask cond and no mask loss & 0.00017 & 0.0024 & 2.892 \\ 
        No mask cond & 0.00012 & 0.0025 & 2.848  \\ 
        No mask loss & 0.00016 & 0.0026 & 3.438  \\ 
        \hline
      \end{tabular}

    \caption{Ablation results for varying the facial semantic information used in the architecture.}
    \label{tab:ablation_sem}
  \end{table}

\subsection{Downstream experiments}


\begin{figure}[t!]
    \centering
    \includegraphics[width=0.9\linewidth]{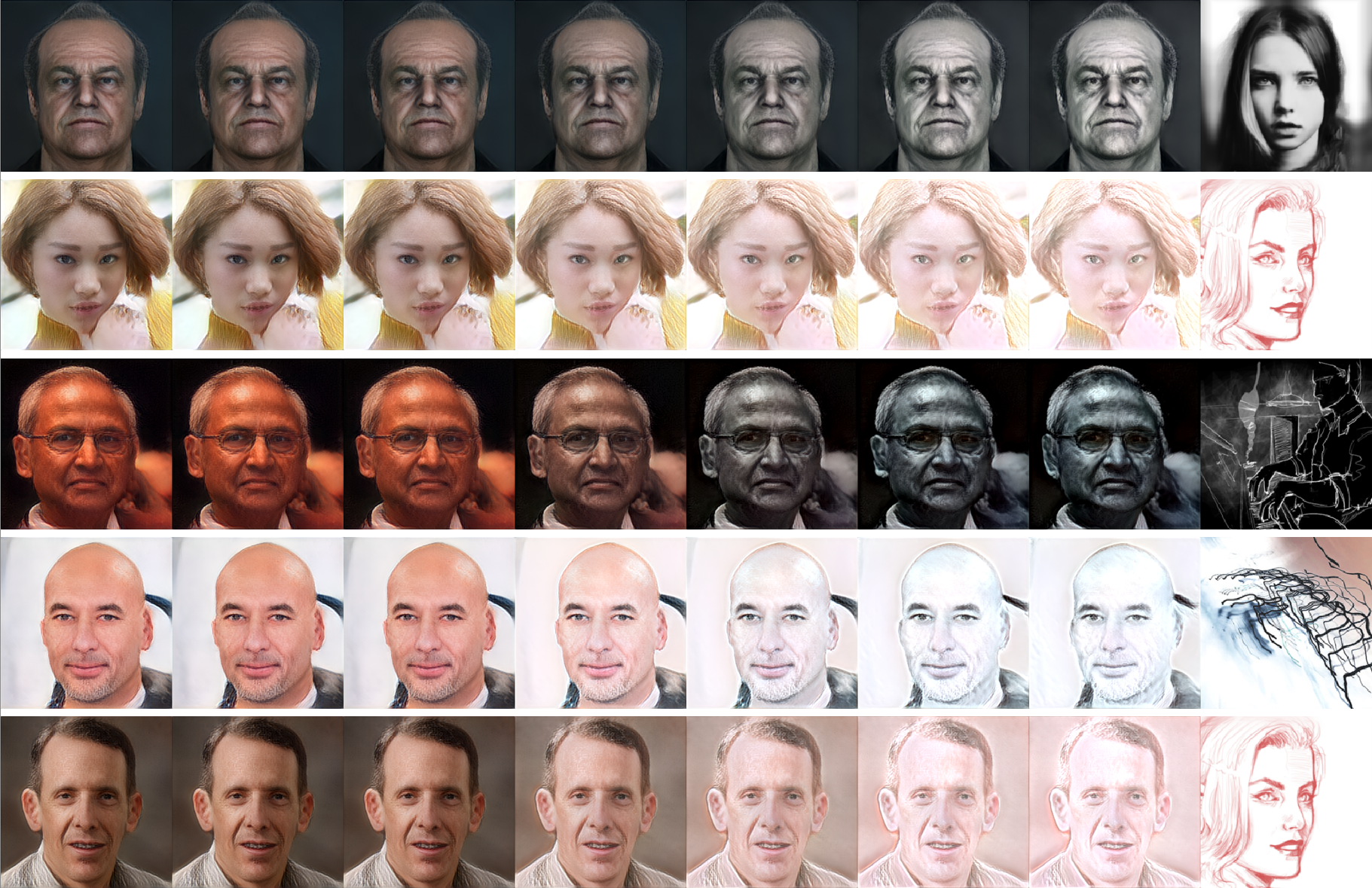} 
    \caption{Visualization of style interpolation between a content image, and a target style image. Note in this example, we show that a non-portrait style image can also be used.}
    \label{fig:interp}
    \squeezeup
\end{figure}

\subsubsection{Style interpolation}

In Figure \ref{fig:interp}, we visualize a controllability aspect of the stylization process based on interpolations done in ALADIN style space. Basing the stylization upon a metric style embedding space, it is possible to perform smooth transitions between the style of a reference content image and a target style image. We also show in this example that it is possible to perform stylization using style images not containing portraits. In this case, the image-wide ALADIN code is used instead of semantically arranged per-feature ALADIN codes.

\begin{figure}[t!]
    \centering
    \includegraphics[width=0.9\linewidth]{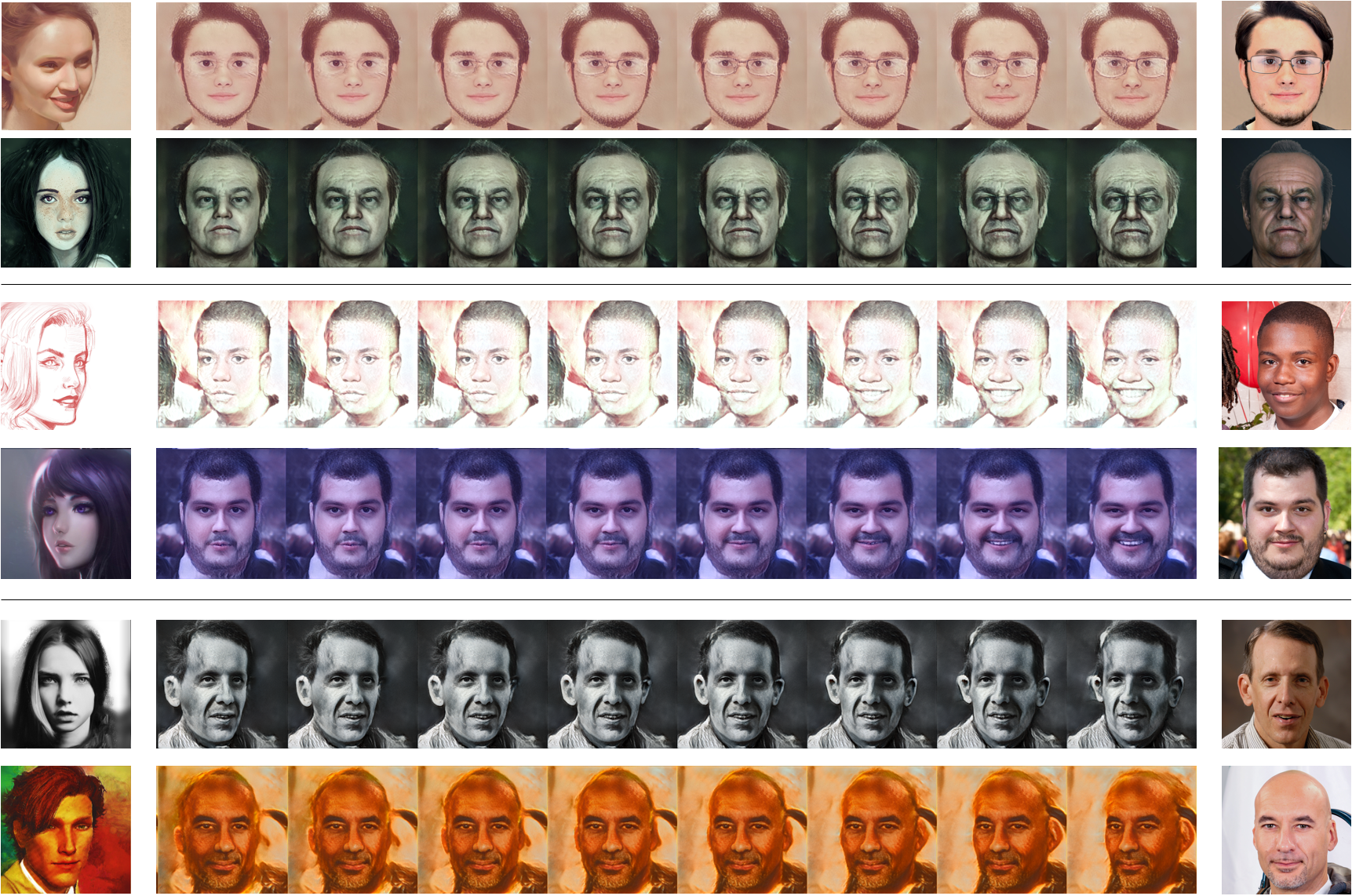} 
    \caption{Semantic face/portrait editing capabilities. The top two example rows range across the \textit{age} vector, the second two example rows show the \textit{smile} vector, and the third shows the \textit{pose} rotation vector.}
    \label{fig:editing}
    \squeezeup
\end{figure}



\subsubsection{Semantic controlability}

One advantage over the SAE model is the editable method of GAN inversion that the HyperNST model uses. Coupled with InterFaceGAN \cite{interfacegan}, latent directions in the StyleGAN2 weight space can be used to expose semantic controls over portrait aspects such as age, pose, and facial expression. On top of the artistic style aspects of the style transfer capabilities of HyperNST, these semantic abilities offer a deeper level of artistic control over the editing and generation of portraits. Figure \ref{fig:editing} visualizes some style transfer examples, where these latent directions are also used in tandem for more advanced control. We show the \textit{smile}, \textit{pose}, and \textit{age} vectors.


\section{Conclusion}

We explored a novel application of a class of models, hypernetworks, to the task of style transfer. We explored the merits and drawbacks of such an approach and what is currently possible. The hypernetwork approach is excellent at maintaining high quality semantic structure and allows for artistic controls, such as semantic editing and style-code-based stylization interpolation. 

The main limitation of the current implementation of hyper-weight prediction is the very high VRAM consumption of such a model (22GB for a batch size of 1) during training, forcing weight deltas to affect the target generator only on a per-channel mean shifting basis.  It is possible that the style transfer quality could be pushed further with a larger batch size. For example, the batch size used by Swapping Autoencoders is 16.  This could be mitigated by exploring smaller alternatives to StyleGAN2, or by implementing GPU optimization tricks such as memory/compute tradeoff training approaches based on gradient checkpointing. Such reductions in VRAM use could also afford an increase in batch size, bringing the training more in line with previous works, primarily affecting discriminators.  A more promising direction could be to more deeply analyse  which layers can be omitted from training (i.e. need  not be updated by the hyper network).   In this paper, we focused our attention on the domain of human portraits. However, other domains would be equally important to study, further generalizing this novel approach to neural artistic stylization using more general semantic segmentation techniques or larger semantically diverse models such as StyleGAN XL, once hardware evolves to support hyper-network inference of larger models' weights.

{\small
\bibliographystyle{ieee_fullname}
\bibliography{main}
}

\end{document}


\pagestyle{headings}
\mainmatter
\def\ECCVSubNumber{4976}  

\title{StyleBabel: Artistic Style Tagging and Captioning} 

\titlerunning{ECCV-22 submission ID \ECCVSubNumber} 
\authorrunning{ECCV-22 submission ID \ECCVSubNumber} 
\author{Supplementary material}
\institute{Paper ID \ECCVSubNumber}



\maketitle
\ificcvfinal\thispagestyle{empty}\fi

\begin{figure}
    \begin{center}
        \includegraphics[width=1\linewidth]{figures/explain_me_3.png}  \\
    \end{center}
    \squeezeup
    \caption{StyleBabel generated caption focuses on style (eg '.. the image has dark tones and high contrast..'), rather than semantics, eg '..a wooded landscape with horse drawn carriage..' from SemArt ground truth}
\end{figure}

\section{StyleBabel Grounded Annotation: Further Information}

Screen captures of the annotation tasks (Stages 1a, 1b, 2 and 3) are presented with representative data fields completed in Figures.~\ref{fig:p1}-\ref{fig:p4} respectively.  Please refer to the main paper for a full account of the Grounded Annotation method used to create StyleBabel .

Briefly, moodboards are annotated by in a pair-wise (contrastive) manner the initial Stage 1a (Fig. \ref{fig:p1}), where an initial tag set (vocabularity) is gathered from individual participants. Each moodboard is annotated by several participants, and the annotation passed into Stage 1b (Fig. \ref{fig:p2}), where the tags are refined as a participant group activity. Once finished, the tags are verified by the ESP-style game in Stage 2 (Fig.\ref{fig:p3}), where any invalid tags are also filtered out.  Descriptive sentences are gathered at this stage. Finally, to refine tag annotation to individual images, these are passed to crowd Stage 3 (Fig.\ref{fig:p4}), where tags invalid for individual images are filtered out through majority consensus.


\begin{figure}
    \begin{center}
        \includegraphics[width=\linewidth]{figures/tasks_screenshots/part1.png}  \\
    \end{center}
    \squeezeup
    \caption{Stage 1a: Pair-wise annotation of moodboards as an individual task. Participants annotate a pair of moodboards with style attributes that are shared by, or distinguish the moodboards.}
    \label{fig:p1}
\end{figure}
\begin{figure}
    \begin{center}
        \includegraphics[width=\linewidth]{figures/tasks_screenshots/part2.png}  \\
    \end{center}
    \squeezeup
    \caption{Stage 1b: Group annotation task to refine annotation and co-create a shared vocabulary among the participant group. Several moodboards are shown to a group of workers with pre-populated tags from Stage 1a. Workers spend roughly one minute per moodboard, making changes to the \textit{stickers} containing tags, after an initial discussion phase.}
    \label{fig:p2}
\end{figure}
\begin{figure}
    \begin{center}
        \includegraphics[width=\linewidth]{figures/tasks_screenshots/part3.png}  \\
    \end{center}
    \squeezeup
    \caption{Stage 2: individual evaluation task to verify the correctness of the tags, as well as generating natural language sentences from the tags.}
    \label{fig:p3}
\end{figure}
\begin{figure}
    \begin{center}
        \includegraphics[width=\linewidth]{figures/tasks_screenshots/part4.png}  \\
    \end{center}
    \squeezeup
    \caption{Stage 3: crowd annotation task where non-experts are shown individual images from the moodboards along with the moodboard tags. They are asked to click which tags do not match the style of the image, to filter out any tag which apply to the moodboard generally, but not to a particular image inside it.}
    \label{fig:p4}
\end{figure}

\section{Data examples}
Tbls. \ref{tab:data_examples_sup1} and \ref{tab:data_examples_sup2} show some additional samples from the StyleBabel dataset. We also include a short sample of the dataset in the companion .csv file, containing image, tags, and caption data.

\begin{figure}
    \begin{center}
        \includegraphics[width=1\linewidth,height=0.5\linewidth]{figures/dataset_tags_distribution.png}  \\
    \end{center}
    \caption{Visualizing occurence of the top tags in StyleBabel.}
\end{figure}

\begin{table*}
  \centering
  \small
      \centering
      \begin{adjustbox}{width=1\textwidth}
          \begin{tabular}{p{0.1\linewidth}|p{0.2\linewidth}|p{0.25\linewidth}|p{0.15\linewidth}|p{0.2\linewidth}|p{0.18\linewidth}}
          
          Image & 
          \vspace{-0.2cm} \raisebox{-\totalheight}{\includegraphics[width=0.15\textwidth]{figures/sup_mat/sup_mat_dataexamples/179f3984520163.5d5f232680f1c.png}} & 
          \vspace{-0.2cm} \raisebox{-\totalheight}{\includegraphics[width=0.15\textwidth]{figures/sup_mat/sup_mat_dataexamples/1106f191201277.5e2b3b0bb2ce5.jpg}} & 
            \vspace{-0.2cm} \raisebox{-\totalheight}{\includegraphics[width=0.15\textwidth]{figures/sup_mat/sup_mat_dataexamples/179d7992738569.5e53505190bed.png}} &   
          \vspace{-0.2cm} \raisebox{-\totalheight}{\includegraphics[width=0.15\textwidth]{figures/sup_mat/sup_mat_dataexamples/25c8af84670353.5d649102e5ec8.png}} & 
          \vspace{-0.2cm} \raisebox{-\totalheight}{\includegraphics[width=0.15\textwidth]{figures/sup_mat/sup_mat_dataexamples/2e2dc573297023.5db78bbe0c4bc.jpg}} \\
          
          \hline
          Tags &
          abstract, portrait, digital, vibrant, texture, painting, up, warm, close, illustration, colorful, bold & 
          abstract, chaotic, complex, texture, colors, pastel, sparse, scale, large, colorful, expressionism & 
          dated, random, promotional, advertising, poster, typography, ux, graphic & 
          line, fantasy, cartoon, bold, black, sketchbook, drawing, story, white, fan, monotone, mystical, art, character, illustration & 
          old, traditional, worn, western, digital, paper, advertisement, stained, cowboy, warm, child, poster, typography, wild, character, graphic  \\
          
          \hline
          Caption & 
          Portrait illustration featuring vibrant colors against a plain background. The digitally rendered painting gives a bold feeling. & 
          Abstract oil painting with textures and splashes is composed of paintbrush strokes. The expressionism artwork is complex and chaotic. & 
          Digital poster illustration featuring bold typography and vector images. The graphic designs in vibrant colors give a dated feel to the promotional material. & 
          Sketch art of line drawing featuring decorative patterns and curved lines against a plain background. The illustration use thin linear and repetitive patterns. &  
          Digital advertisement poster layout using retro style with linear line art graphics against a stained background with border. The content consists of digital graphic designs with cluttered informative typography using stylized fonts. \\
          
          \hline
          
            Image &
          \vspace{-0.2cm} \raisebox{-\totalheight}{\includegraphics[width=0.15\textwidth]{figures/sup_mat/sup_mat_dataexamples/0b25a288448625.5dd66beb14a66.jpg}} & 
          \vspace{-0.2cm} \raisebox{-\totalheight}{\includegraphics[width=0.15\textwidth]{figures/sup_mat/sup_mat_dataexamples/02d90f89187751.5ded7326a916a.jpg}} & 
            \vspace{-0.2cm} \raisebox{-\totalheight}{\includegraphics[width=0.15\textwidth]{figures/sup_mat/sup_mat_dataexamples/29ad5887642795.5dbed8fc12caa.jpg}}     &
          \vspace{-0.2cm} \raisebox{-\totalheight}{\includegraphics[width=0.15\textwidth]{figures/sup_mat/sup_mat_dataexamples/1ad2d688893565.5de4d7db412e3.jpg}} & 
          \vspace{-0.2cm} \raisebox{-\totalheight}{\includegraphics[width=0.15\textwidth]{figures/sup_mat/sup_mat_dataexamples/32312681662205.5d37fd0747ee9.jpg}} \\
          
          \hline
          Tags &
          complementary, pattern, bright, crowded, rounded, fluid, motif, doodle, monochrome, sporadic, repetitive, graphic & 
          abandoned, dark, derelict, dull, ghostly, spooky & 
          logo, identity, clean, outline, thick, animal, black, visual, branding, white, iconography, type, monochrome, vector, icon, typography & 
          360, industrial, photography, aerial, natural, conceptual, architectural, drone, render, dull, landscape & 
          abstract, chaotic, complex, texture, colors, pastel, sparse, scale, large, colorful, expressionism
           \\
          \hline
          Caption &
          Layout design featuring a graphical design with a crowded and repetitive style. The design contains small motifs and doodles in different shapes with a vibrant and bright style against the white background. & 
          Wide-angle photography using natural warm lighting and shadow effect of an abandoned architectural structure with cylindrical shapes and grids. The dramatic photograph captures the golden hue and foamy wave pattern in nature. & 
          Digital typographic logo design using graphic elements and stylized font against a dark background. The color scheme is contrasting. & 
          3d rendered mock-up of architectural design from an aerial shot. The conceptual layout shows industrial structures with symmetrical shapes and designs against a landscape background. & 
          Abstract oil painting featuring chaotic and complex linear patterns. This expressionism also features sparse textures using pastel and muted colors.
          \\
                     
            \end{tabular}
      \end{adjustbox}
    \caption{Excerpt of StyleBabel dataset. Images and the corresponding tags and sentences collected via our Grounded Annotation.}
    \label{tab:data_examples_sup1}
    \vspace{10cm}
\end{table*}         


\begin{table*}
  \centering
  \small
      \centering
      \begin{adjustbox}{width=1\textwidth}
          \begin{tabular}{p{0.1\linewidth}|p{0.25\linewidth}|p{0.2\linewidth}|p{0.2\linewidth}|p{0.2\linewidth}|p{0.18\linewidth}}
          
            Image &
          \vspace{-0.2cm} \raisebox{-\totalheight}{\includegraphics[width=0.15\textwidth]{figures/sup_mat/sup_mat_dataexamples/32318876288699.5e209dbe8895c.png}} & 
          \vspace{-0.2cm} \raisebox{-\totalheight}{\includegraphics[width=0.15\textwidth]{figures/sup_mat/sup_mat_dataexamples/1d322089772409.5e002630bca5b.jpg}} & 
          
          \vspace{-0.2cm} \raisebox{-\totalheight}{\includegraphics[width=0.15\textwidth]{figures/sup_mat/sup_mat_dataexamples/3a48ac87478815.5db97943cc3b9.png}} & 
          \vspace{-0.2cm} \raisebox{-\totalheight}{\includegraphics[width=0.15\textwidth]{figures/sup_mat/sup_mat_dataexamples/021b4784519447.5d5ffcab54985.jpg}} & 
          \vspace{-0.2cm} \raisebox{-\totalheight}{\includegraphics[width=0.15\textwidth]{figures/sup_mat/sup_mat_dataexamples/11054386230395.5d9350e3ab18e.jpg}} \\
          
          \hline
          Tags &
          business, logotype, identity, personal, serif, mock-up, simple, sleek, visual, branding, minimal, sans, promotional, professional, monochrome, formal, font, card & 
          creative, graphic, paper, animated, flat, development, sketch, drawing, artistic, amateur, professional, illustration, animation, character, storyboard & 
          creative, graphic, paper, flat, animated, development, sketch, drawing, artistic, amateur, professional, illustration, animation, character, storyboard & 
          mock-up, black, simple, branding, white, classy, stylish, book & 
          contrast, kaleidoscope, psychedelic, complex, intricate, complicated, paisley, colored, colorful, repetitive, surreal
           \\
          \hline
          Caption &
          Mock-up of formal business cards with simple and sleek typography using serif font. This visual identity layout also features a logo design with green outlining displayed on a plain background. & 
          Sketch illustration featuring creative graphic design. Flat animated drawing is detailed with artistic lines for character development. & 
          Raw digital sketch representing an artistic storyboard on development stage with structured character design and animated setup formation. The creative drawing has an amateur approach with flat visual presentation. & 
          3D rendered book mock-up for publication. The simple monochromatic cover page looks classy, featuring graphic design, cropped photograph of a female and minimal typography against a black background. &
          kaleidoscopic illustration featuring repetitive psychedelic pattern with abstract shapes against a plain background. The colorful artwork is complex and intricate.\\
        \hline     
            Image &
          \vspace{-0.2cm} \raisebox{-\totalheight}{\includegraphics[width=0.15\textwidth]{figures/sup_mat/sup_mat_dataexamples/39a1c082485587.5d4fd9ccb927a.jpg}} & 
          \vspace{-0.2cm} \raisebox{-\totalheight}{\includegraphics[width=0.15\textwidth]{figures/sup_mat/sup_mat_dataexamples/140dea92900207.5e57051f076ea.jpg}} & 
          \vspace{-0.2cm} \raisebox{-\totalheight}{\includegraphics[width=0.15\textwidth]{figures/sup_mat/sup_mat_dataexamples/213d2283966517.5d4c7e33bdec4.jpg}} & 
          \vspace{-0.2cm} \raisebox{-\totalheight}{\includegraphics[width=0.15\textwidth]{figures/sup_mat/sup_mat_dataexamples/140e2090451123.5e1775a70eaad.jpg}} & 
          \vspace{-0.2cm} \raisebox{-\totalheight}{\includegraphics[width=0.15\textwidth]{figures/sup_mat/sup_mat_dataexamples/0232eb82652111.5d2454d720857.jpg}} \\
          
          \hline
          Tags &
          industrial, clean, structured, bright, architectural, masculine, modern, futuristic, linear, monotone, mechanical & 
          fine, texture, pattern, colors, detailed, geometric, secondary, repetitive & 
          interior, concept, retail, model, architectural, mock-up, prototype, commercial, 3d, render, floorplan, magazine, generation, section, editorial  & 
          abstract, paint, watercolor, psychedelic, fine, texture, pattern, oil, colors, floral, crowded, warm, vivid, art, overwhelming, colorful, repetitive & 
          dark, contrast, digital, vibrant, bright, colors, dynamic, manipulation, neon, fan, illustration, art, object \\
          \hline
          Caption &
          Photography of a luxurious interior design with structured grids and artificial light. The formation of the multifunctional industrial space is symmetrically aligned and looks clean. & 
          Illustration featuring repetitive patterns and fine textures against a plain background. The detailed artwork is composed of geometric shapes and secondary tones which creates an illusionary visual effect. & 
          Digital layout, presenting a centered conceptual 3D architectural interior floor-planning design developed in linear, and textured patterns in warm neutral tones. The prototype is displayed against a plain background. & 
          Abstract watercolor painting featuring repetitive psychedelic patterns and floral design. The colorful crowded artwork is overwhelming with the composition of fine textures and shades. & 
          Digitally manipulated illustration featuring a futuristic theme, a photograph with perspective distortion, geometric shapes and ripple and dot patterns. The dynamic image is eye catching with vibrant neon colors against a dark background. \\
           
          \end{tabular}
      \end{adjustbox}
    \caption{Excerpt of StyleBabel dataset.  Images and the corresponding tags and sentences collected via our Grounded Annotation.}
    \label{tab:data_examples_sup2}
\end{table*}

\section{Model generated results}
In Fig.\ref{fig:senteneces_viz} we show captions generated using our VirTex model trained with MS-COCO (top) and StyleBabel (bottom). We present some further examples (in Figs. \ref{fig:tags_viz1} and \ref{fig:tags_viz2}) of tags generated using our ALADIN-ViT style embeddings and our CLIP model. Fig \ref{fig:word_dist} shows word type distribution.


\begin{figure*}
  \centering
  \small
      \centering
      \begin{adjustbox}{width=1\textwidth}
          \begin{tabular}{p{0.12\linewidth}p{0.21\linewidth}p{0.20\linewidth}p{0.20\linewidth}p{0.20\linewidth}p{0.20\linewidth} }

            &
            \vspace{-0.2cm} \raisebox{-\totalheight}{\includegraphics[width=0.20\textwidth]{figures/sup_mat/3c34c890554585.5e1a905405027.jpg}} & 
            \vspace{-0.2cm} \raisebox{-\totalheight}{\includegraphics[width=0.20\textwidth]{figures/sup_mat/3a87a790970989.5e25e7a7971f3.jpg}} & 
            \vspace{-0.2cm} \raisebox{-\totalheight}{\includegraphics[width=0.20\textwidth]{figures/sup_mat/4c7a9990308405.5e144baad712e.jpg}} & 
            \vspace{-0.2cm} \raisebox{-\totalheight}{\includegraphics[width=0.20\textwidth]{figures/sup_mat/4fc30a88811537.5de1703f1fc2c.jpg}} & 
            \vspace{-0.2cm} \raisebox{-\totalheight}{\includegraphics[width=0.20\textwidth]{figures/sup_mat/6d34c285884191.5d896be050f03.png}} \\

            MS-COCO &
            a close up of a bird in a tree &
            a bunch of flowers are in a vase &
            a bottle of beer sitting on top of a red & 
            a colorful building with a painted painted on it & 
            a picture of a picture of a picture of a wall \\
            
            \hline
            StyleBabel CL+IL &
            the pencil sketch is a soft linear drawing with the use of shades and strokes against a plain background & 
            digital illustration of a fictional character using a dark color palette & 
            digital illustration featuring repetitive shapes and patterns using a color palette of blue and red & 
            abstract painting with geometric shapes and patterns using bright colors & 
            digital illustration of a fictional character using a color palette of black and white 
            
            \\
            &
            \vspace{-0.2cm} \raisebox{-\totalheight}{\includegraphics[width=0.20\textwidth]{figures/sup_mat/3a605989034299.5de84faf582c3.jpg}} & 
            \vspace{-0.2cm} \raisebox{-\totalheight}{\includegraphics[width=0.20\textwidth]{figures/sup_mat/08ab3e92745943.5e53931d331d3.jpg}} & 
            \vspace{-0.2cm} \raisebox{-\totalheight}{\includegraphics[width=0.20\textwidth]{figures/sup_mat/5a463d78184135.5d636fbe313fe.jpg}} & 
            \vspace{-0.2cm} \raisebox{-\totalheight}{\includegraphics[width=0.20\textwidth]{figures/sup_mat/24995f82254331.5d17eb08468fc.jpg}} & 
            \vspace{-0.2cm} \raisebox{-\totalheight}{\includegraphics[width=0.20\textwidth]{figures/sup_mat/70a30490377237.5e15fb4291611.jpg}} \\ 
            
            MS-COCO &
            a group of birds are in the sky &
            a display of a display with many clocks &
            a group of birds that are on a pole & 
            a close up of a picture of the water & 
            a close up of a person with a black background \\ 
            
            \hline
            StyleBabel CL+IL &
            digital layout of a logo with minimal typography and a white background & 
            digital illustration of cartoon characters using bright colors against a blue background & 
            digital illustration of a sketch with line drawings on a white background & 
            abstract oil painting using paintbrush strokes and scribbles using a cool color palette & 
            digital illustration of repetitive linear pattern using blue and white colors

            \\
            &
            \vspace{-0.2cm} \raisebox{-\totalheight}{\includegraphics[width=0.20\textwidth]{figures/sup_mat/3313e386308243.5d955422085fd.jpg}} & 
            \vspace{-0.2cm} \raisebox{-\totalheight}{\includegraphics[width=0.20\textwidth]{figures/sup_mat/4e630e88522319.5dd8754bb267c.jpg}} & 
            \vspace{-0.2cm} \raisebox{-\totalheight}{\includegraphics[width=0.20\textwidth]{figures/sup_mat/2af18c88707049.5dde6750208bd.jpg}} & 
            \vspace{-0.2cm} \raisebox{-\totalheight}{\includegraphics[width=0.20\textwidth]{figures/sup_mat/80847787607963.5dbd617e4b96b.jpg}} & 
            \vspace{-0.2cm} \raisebox{-\totalheight}{\includegraphics[width=0.20\textwidth]{figures/sup_mat/6dc7a290265653.5e133548276c7.jpg}} \\ 
            
            MS-COCO &
            a picture of a wall of a wall with a wall &
            a close up of a woman with a hair &
            a colorful decorated decorated decorated with colorful decorations & 
            a man in a red jacket is standing in front of a store & 
            a white and white building with a white and white building \\ 
            
            \hline
            StyleBabel CL+IL &
            digital illustration of a 3d rendered illustration with repetitive shapes and patterns & 
            the sketch is a soft linear drawing with the use of shades and strokes against a white background & 
            digital illustration of an animated character using bright colors & 
            event photography featuring subject in center with colorful lighting against a blue background & 
            3d rendered modern interior design with repetitive straight lines and shapes

          \end{tabular}
      \end{adjustbox}
      \squeezeup
    \caption{Examples of natural language sentences generated for various art styles; The top row shows captions generated using a VirTex model trained on MS-COCO, and the bottom row shows captions generated using VirTex trained on StyleBabel.}
    \label{fig:senteneces_viz}
    \squeezeup
\end{figure*}

\begin{figure*}
  \centering
  \small
      \centering
      \begin{adjustbox}{width=\textwidth}
          \begin{tabular}{p{0.20\linewidth}p{0.20\linewidth}p{0.20\linewidth}p{0.20\linewidth}p{0.20\linewidth} }

            \vspace{-0.2cm} \raisebox{-\totalheight}{\includegraphics[width=0.20\textwidth]{figures/img_to_tag/Picture15.png}} & \vspace{-0.2cm} \raisebox{-\totalheight}{\includegraphics[width=0.20\textwidth]{figures/img_to_tag/Picture16.png}} & \vspace{-0.2cm} \raisebox{-\totalheight}{\includegraphics[width=0.20\textwidth]{figures/img_to_tag/Picture17.png}} & \vspace{-0.2cm} \raisebox{-\totalheight}{\includegraphics[width=0.20\textwidth]{figures/img_to_tag/Picture18.png}} & \vspace{-0.2cm} \raisebox{-\totalheight}{\includegraphics[width=0.20\textwidth]{figures/img_to_tag/Picture19.png}} \\ 
            sketch, claustrophobic, scamp, line, observational & 
            paint, abstract, brush, fine, mark-making & 
            flick, rendition, generative, line, curvature & 
            branding, cu, typography, logo, typeface & 
            collage, art, nonobjective, basquiat, illustration \\
            
            \vspace{-0.2cm} \raisebox{-\totalheight}{\includegraphics[width=0.20\textwidth]{figures/img_to_tag/Picture19.png}} & \vspace{-0.2cm} \raisebox{-\totalheight}{\includegraphics[width=0.20\textwidth]{figures/img_to_tag/Picture20.png}} & \vspace{-0.2cm} \raisebox{-\totalheight}{\includegraphics[width=0.20\textwidth]{figures/img_to_tag/Picture21.png}} & \vspace{-0.2cm} \raisebox{-\totalheight}{\includegraphics[width=0.20\textwidth]{figures/img_to_tag/Picture22.png}} & \vspace{-0.2cm} \raisebox{-\totalheight}{\includegraphics[width=0.20\textwidth]{figures/img_to_tag/Picture23.png}} \\ 
            collage, art, nonobjective, basquiat, illustration & 
            branding, logo, brief, cu, icon  & 
            expressive, painted, oil, full, fairy & 
            dynamic, imaginary, fantasy, illustration, villain & 
            cotton, dress, hanging, fabric, fashion \\

            \vspace{-0.2cm} \raisebox{-\totalheight}{\includegraphics[width=0.20\textwidth]{figures/img_to_tag/Picture24.jpg}} & \vspace{-0.2cm} \raisebox{-\totalheight}{\includegraphics[width=0.20\textwidth]{figures/img_to_tag/Picture25.jpg}} & \vspace{-0.2cm} \raisebox{-\totalheight}{\includegraphics[width=0.20\textwidth]{figures/img_to_tag/Picture26.jpg}} & \vspace{-0.2cm} \raisebox{-\totalheight}{\includegraphics[width=0.20\textwidth]{figures/img_to_tag/Picture27.jpg}} & \vspace{-0.2cm} \raisebox{-\totalheight}{\includegraphics[width=0.20\textwidth]{figures/img_to_tag/Picture28.jpg}} \\ 
            corporate, commercial, business, formal, visual & 
            wavy, graffiti, mural, tagging, colorful & 
            cartoon, character, playful, fun, pop & 
            logo, logotype, branding, brief, icon & 
            web, ux, ui, corporate, common \\            
            
            \vspace{-0.2cm} \raisebox{-\totalheight}{\includegraphics[width=0.20\textwidth]{figures/img_to_tag/Picture29.png}} & \vspace{-0.2cm} \raisebox{-\totalheight}{\includegraphics[width=0.20\textwidth]{figures/img_to_tag/Picture30.png}} & \vspace{-0.2cm} \raisebox{-\totalheight}{\includegraphics[width=0.20\textwidth]{figures/img_to_tag/Picture31.png}} & \vspace{-0.2cm} \raisebox{-\totalheight}{\includegraphics[width=0.20\textwidth]{figures/img_to_tag/Picture32.png}} & \vspace{-0.2cm} \raisebox{-\totalheight}{\includegraphics[width=0.20\textwidth]{figures/img_to_tag/Picture33.png}} \\ 
            vector, character, punk, digital, caricature & 
            hand-rendered, rough, suburban, outline, line & 
            caricature, character, cartoon, flat, vector & 
            line, generative, hand-drawn, drawn, stylized & 
            brand, embossed, branding, identity, visualization \\
                        
            \vspace{-0.2cm} \raisebox{-\totalheight}{\includegraphics[width=0.20\textwidth]{figures/img_to_tag/Picture34.png}} & \vspace{-0.2cm} \raisebox{-\totalheight}{\includegraphics[width=0.20\textwidth]{figures/img_to_tag/Picture35.png}} & \vspace{-0.2cm} \raisebox{-\totalheight}{\includegraphics[width=0.20\textwidth]{figures/img_to_tag/Picture36.png}} & \vspace{-0.2cm} \raisebox{-\totalheight}{\includegraphics[width=0.20\textwidth]{figures/img_to_tag/Picture37.png}} & \vspace{-0.2cm} \raisebox{-\totalheight}{\includegraphics[width=0.20\textwidth]{figures/img_to_tag/Picture38.png}} \\ 
            paint, abstract, oil, colorful, flow & 
            advertising, bright, garish, marketing, rainbow & 
            prototype, mock-up, information, proposal, promotional & 
            creative, imaginative, pop, graphic, illustration & 
            fashion, studio, model, designer, magazine \\

          \end{tabular}
      \end{adjustbox}
      \vspace{0.3cm}
    \caption{Style2Text tag generation experiment, using CLIP trained with ALADIN-ViT encoder. Top 5 tags shown for each image.}
    \label{fig:tags_viz1}
\end{figure*}

\begin{figure*}
  \centering
  \small
      \centering
      \begin{adjustbox}{width=\textwidth}
          \begin{tabular}{p{0.20\linewidth}p{0.20\linewidth}p{0.20\linewidth}p{0.20\linewidth}p{0.20\linewidth} }
          
          \vspace{-0.2cm} \raisebox{-\totalheight}{\includegraphics[width=0.20\textwidth]{figures/img_to_tag/Picture39.png}} & \vspace{-0.2cm} \raisebox{-\totalheight}{\includegraphics[width=0.20\textwidth]{figures/img_to_tag/Picture40.png}} & \vspace{-0.2cm} \raisebox{-\totalheight}{\includegraphics[width=0.20\textwidth]{figures/img_to_tag/Picture41.png}} & \vspace{-0.2cm} \raisebox{-\totalheight}{\includegraphics[width=0.20\textwidth]{figures/img_to_tag/Picture42.png}} & \vspace{-0.2cm} \raisebox{-\totalheight}{\includegraphics[width=0.20\textwidth]{figures/img_to_tag/Picture43.png}} \\ 
            generative, lined, artisan, complex, pattern & 
            brand, sleek, branding, profession, printing & 
            paint, abstract, particle, whimsical, oil & 
            cartoon, illustration, comic, trip, storyboard & 
            nerdy, black, generative, monochrome, dark \\
                        
            \vspace{-0.2cm} \raisebox{-\totalheight}{\includegraphics[width=0.20\textwidth]{figures/img_to_tag/Picture44.png}} & \vspace{-0.2cm} \raisebox{-\totalheight}{\includegraphics[width=0.20\textwidth]{figures/img_to_tag/Picture45.png}} & \vspace{-0.2cm} \raisebox{-\totalheight}{\includegraphics[width=0.20\textwidth]{figures/img_to_tag/Picture46.png}} & \vspace{-0.2cm} \raisebox{-\totalheight}{\includegraphics[width=0.20\textwidth]{figures/img_to_tag/Picture47.png}} & \vspace{-0.2cm} \raisebox{-\totalheight}{\includegraphics[width=0.20\textwidth]{figures/img_to_tag/Picture48.png}} \\ 
            lightened, cozy, interior, homely, luxurious & 
            distorted, creative, bold, illustration, nonobjective & 
            mark, icon, vector, monogram, logo & 
            freehand, line, stroke, thin, freeform & 
            cartoon, character, avatar, child, animation \\
            
            \vspace{-0.2cm} \raisebox{-\totalheight}{\includegraphics[width=0.20\textwidth]{figures/img_to_tag/Picture49.png}} & \vspace{-0.2cm} \raisebox{-\totalheight}{\includegraphics[width=0.20\textwidth]{figures/img_to_tag/Picture50.png}} & \vspace{-0.2cm} \raisebox{-\totalheight}{\includegraphics[width=0.20\textwidth]{figures/img_to_tag/Picture51.png}} & \vspace{-0.2cm} \raisebox{-\totalheight}{\includegraphics[width=0.20\textwidth]{figures/img_to_tag/Picture52.png}} & \vspace{-0.2cm} \raisebox{-\totalheight}{\includegraphics[width=0.20\textwidth]{figures/img_to_tag/Picture53.png}} \\ 
            informative, copy, complex, presentation, web & 
            sketchbook, original, sketching, line, quick & 
            graffiti, bold, abstract, tagging, mural & 
            headshot, portrait, selfie, hair, expression & 
            centered, dainty, hand-lettering, calligraphy, informal \\     
             
            \vspace{-0.2cm} \raisebox{-\totalheight}{\includegraphics[width=0.20\textwidth]{figures/img_to_tag/Picture54.png}} & \vspace{-0.2cm} \raisebox{-\totalheight}{\includegraphics[width=0.20\textwidth]{figures/img_to_tag/Picture55.png}} & \vspace{-0.2cm} \raisebox{-\totalheight}{\includegraphics[width=0.20\textwidth]{figures/img_to_tag/Picture56.png}} & \vspace{-0.2cm} \raisebox{-\totalheight}{\includegraphics[width=0.20\textwidth]{figures/img_to_tag/Picture57.png}} & \vspace{-0.2cm} \raisebox{-\totalheight}{\includegraphics[width=0.20\textwidth]{figures/img_to_tag/Picture58.png}} \\ 
            typeface, basic, logo, ligature, font & 
            calligraphy, flick, curvature, generative, hand-lettering & 
            interior, fengshui, cozy, cosmopolitan, lightened & 
            fashion, magazine, persona, material, emo & 
            promotion, beverage, upbeat, advertisement, conclusive \\    
             
            \vspace{-0.2cm} \raisebox{-\totalheight}{\includegraphics[width=0.20\textwidth]{figures/img_to_tag/Picture59.png}} & \vspace{-0.2cm} \raisebox{-\totalheight}{\includegraphics[width=0.20\textwidth]{figures/img_to_tag/Picture60.png}} & \vspace{-0.2cm} \raisebox{-\totalheight}{\includegraphics[width=0.20\textwidth]{figures/img_to_tag/Picture61.png}} & \vspace{-0.2cm} \raisebox{-\totalheight}{\includegraphics[width=0.20\textwidth]{figures/img_to_tag/Picture62.png}} & \vspace{-0.2cm} \raisebox{-\totalheight}{\includegraphics[width=0.20\textwidth]{figures/img_to_tag/Picture63.png}} \\ 
            novel, abstract, lined, stencil, hand-rendered & 
            informative, publication, page, academic, readable & 
            abstract, paint, oil, fine, mark-making & 
            rasping, typography, red, eye-catchy, ambiguous & 
            pattern, agglomeration, texture, kaleidoscope, 80 \\

            
          \end{tabular}
      \end{adjustbox}
      \vspace{0.3cm}
    \caption{Style2Text tag generation experiment, using CLIP trained with ALADIN-ViT encoder. Top 5 tags shown for each image.}
    \label{fig:tags_viz2}
\end{figure*}



\begin{figure*}
    \begin{center}
        \includegraphics[width=0.6\linewidth]{figures/POS.png}  \\
    \end{center}
    \caption{Distribution of word types for the set of unique style attributes (tags) collected for  images within the StyleBabel dataset}
    \label{fig:word_dist}
\end{figure*}

\begin{figure*}
    \begin{center}
        \includegraphics[width=1\linewidth]{figures/sup_mat_caption_occurences.png}  \\
    \end{center}
    \caption{Occurences of the top words in the StyleBabel captions}
    \label{fig:word_dist}
\end{figure*}
